\documentclass[10pt,twocolumn,letterpaper]{article}

\usepackage{iccv}
\usepackage{times}
\usepackage{epsfig}
\usepackage{graphicx}
\usepackage{amsmath}
\usepackage{amssymb}

\usepackage{booktabs}
\usepackage{multirow}
\usepackage{multicol}
\usepackage[table]{xcolor}
\usepackage{arydshln}
\usepackage{subcaption}

\usepackage[pagebackref=true,breaklinks=true,letterpaper=true,colorlinks,bookmarks=false]{hyperref}

\iccvfinalcopy 

\def\iccvPaperID{9809} 

\ificcvfinal\fi

\begin{document}

\title{ParCNetV2: Oversized Kernel with Enhanced Attention}
\author{Ruihan Xu$^{1}$, Haokui Zhang$^{2,3}$, Wenze Hu$^{2}$, Shiliang Zhang$^{1}$, Xiaoyu Wang$^{2}$\\
  $^{1}$Peking University, Beijing, China.  
  $^{2}$Intellifusion, Shenzhen, China\\
  $^{3}$Harbin Institute of Technology (Shenzhen), Shenzhen, China\\
}

\maketitle
\ificcvfinal\thispagestyle{empty}\fi

\begin{abstract}
  Transformers have shown great potentials in various computer vision tasks.
  By borrowing design concepts from transformers, many studies revolutionized CNNs and showed remarkable results.
  This paper falls in this line of studies.
  Specifically, we propose a new convolutional neural network, \textbf{ParCNetV2}, that extends position-aware circular convolution (ParCNet) with oversized convolutions and bifurcate gate units to enhance attention.
  The oversized convolution employs a kernel with twice the input size to model long-range dependencies through a global receptive field.
  Simultaneously, it achieves implicit positional encoding by removing the shift-invariant property from convolution kernels, \emph{i.e.}, the effective kernels at different spatial locations are different when the kernel size is twice as large as the input size.
  The bifurcate gate unit implements an attention mechanism similar to self-attention in transformers.
  It is applied through element-wise multiplication of the two branches, one serves as feature transformation while the other serves as attention weights.
  Additionally, we introduce a uniform local-global convolution block to unify the design of the early and late stage convolution blocks. 
  Extensive experiments demonstrate the superiority of our method over other convolutional neural networks and hybrid models that combine CNNs and transformers. Code will be released.

\end{abstract}

\section{Introduction}
\label{sec:intro}

\begin{figure}
  \centering
  \includegraphics[width=0.49\linewidth]{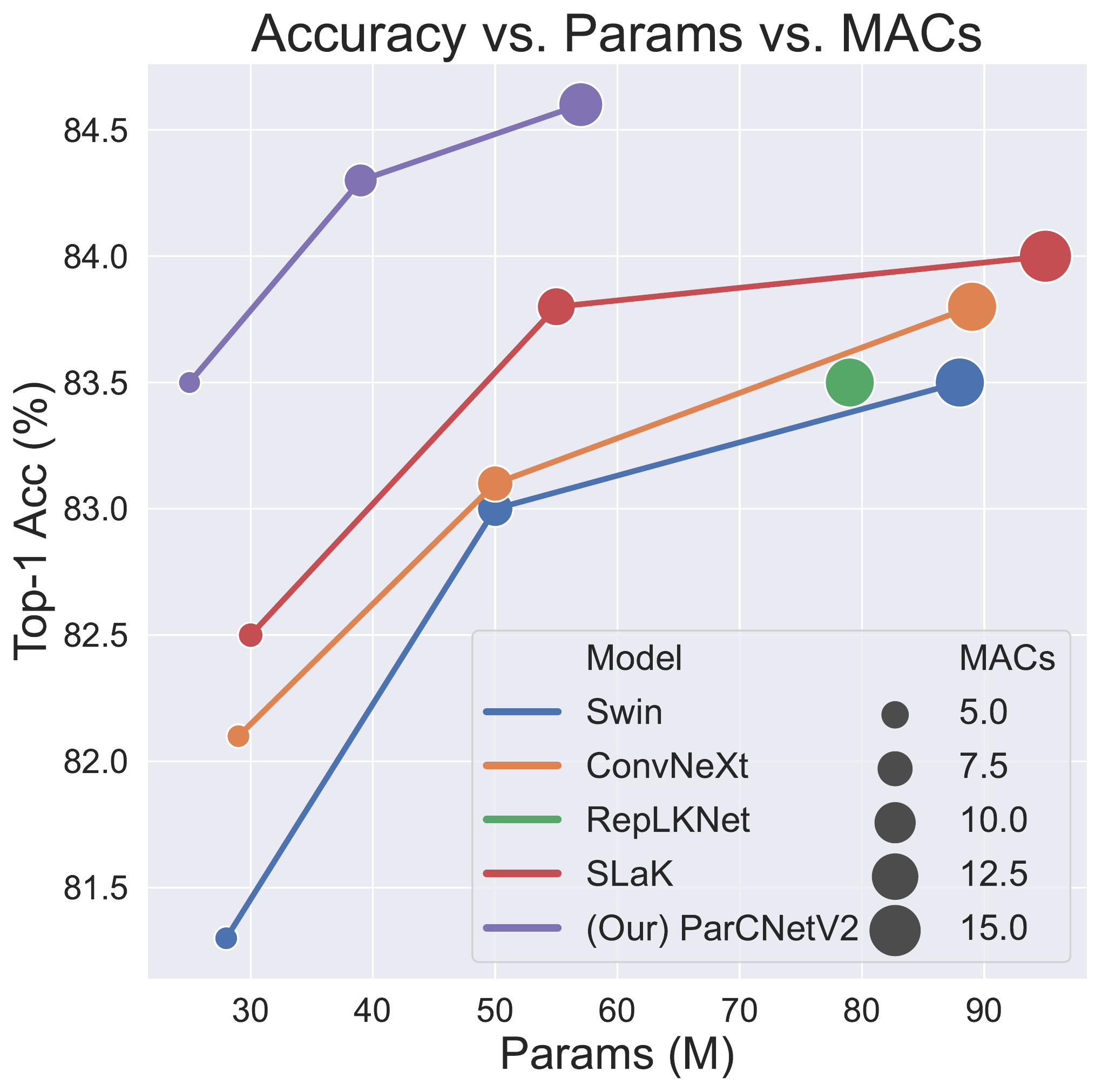}
  \includegraphics[width=0.49\linewidth]{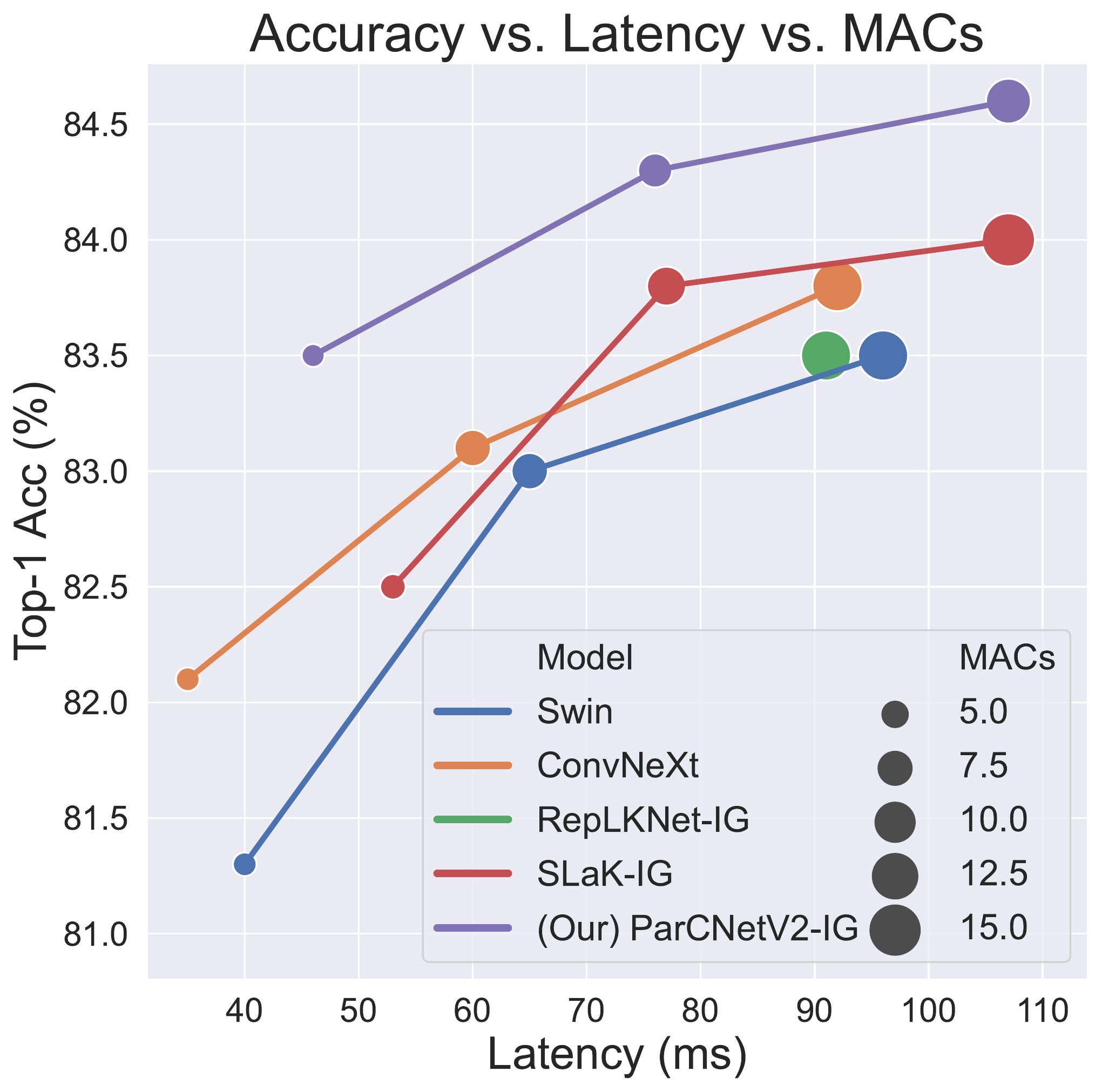}
  \caption{Comparison between ParCNetV2 with the prevailing transformer (Swin), CNN (ConvNeXt), and large kernel CNNs (RepLKNet \& SLaK) when trained from scratch on ImageNet-1K. Left: performance curve of model size vs. top-1 accuracy. Right: performance curve of inference latency vs. top-1 accuracy. \textbf{IG} represents using the \textit{implicit gemm} acceleration algorithm.}
  \label{fig:trade_offs}
  \vspace{-3mm}
\end{figure}

Transformers have shown great potential in computer vision recently. Vision transformer (ViT)~\cite{dosovitskiy2020image} and its variants~\cite{touvron2021training,yuan2021tokens,wang2021pyramid,liu2021swin} have been adopted to various vision tasks such as object detection~\cite{carion2020end,fang2021you}, semantic segmentation~\cite{zheng2021rethinking}, and multi-modal tasks such as visual question answering~\cite{khan2020mmft} and text-to-image synthesis~\cite{ramesh2022hierarchical}. Despite the great performance of vision transformers, they do not win convolutional neural networks (CNNs) in all aspects. For example, the computational complexity of self-attention modules, one of the critical designs in transformers, is quadratic ($\mathcal{O}(N^2C)$) to the resolution of inputs~\cite{vaswani2017attention}. This property restricts its adoption in real applications such as defect inspection, which finds small defects in high-resolution images~\cite{zhang2021multi}. Moreover, transformers are arguably more data-hungry than CNNs~\cite{dosovitskiy2020image,touvron2021training}, making them difficult to be deployed to long-tail applications where no large-scale data is available. Lastly, CNNs have been intensively studied in the past several decades~\cite{lecun1995convolutional}. There are lots of off-the-shelf dedicated features already developed in existing deployment hardware (CPU, GPU, FPGA, ASIC, \emph{etc.}). Some acceleration and deployment techniques are designed mainly around convolution operations, such as operator fusion~\cite{roesch2019relay} and multi-level tiling~\cite{zheng2020ansor,chen2018tvm}.

Thus pushing the envelope of CNNs is still important and valuable. Recent works have improved CNNs from multiple perspectives. A straightforward approach is to take the benefits from both CNNs and transformers by mixing their building blocks ~\cite{graham2021levit,srinivas2021bottleneck,mehta2021mobilevit,chen2022mobile,li2022efficientformer}. While bringing together merits from the two parties, those approaches still keep the ViT blocks and has the quadratic complexity problem. Another line of research is to design purely convolutional architectures. For example, with larger convolution kernels,  ConvNeXt~\cite{liu2022convnet}, RepLKNet~\cite{ding2022scaling}, and ParCNetV1~\cite{zhang2022parc} successfully improved the performance of CNNs by encoding broader spatial contexts.

Specifically, ParCNetV1 introduced \textbf{p}osition-\textbf{a}ware ci\textbf{r}cular  \textbf{c}onvolutions (ParC) to CNNs.
It uses depth-wise circular 1D convolutions of input feature map size ($C \times H \times 1$ and $C \times 1 \times W$) to achieve global receptive fields.
To avoid spatial over-smoothing caused by global kernels, ParCNetV1 augmented the feature input with absolute position encoding to ensure the feature output is still location sensitive.
It also brought attention mechanisms into the framework by adopting squeeze-and-excitation block~\cite{hu2018squeeze}.
These modifications lead to the superior performance of ParCNetV1, especially on mobile devices.

\begin{figure}
  \centering
  \includegraphics[width=0.9\linewidth]{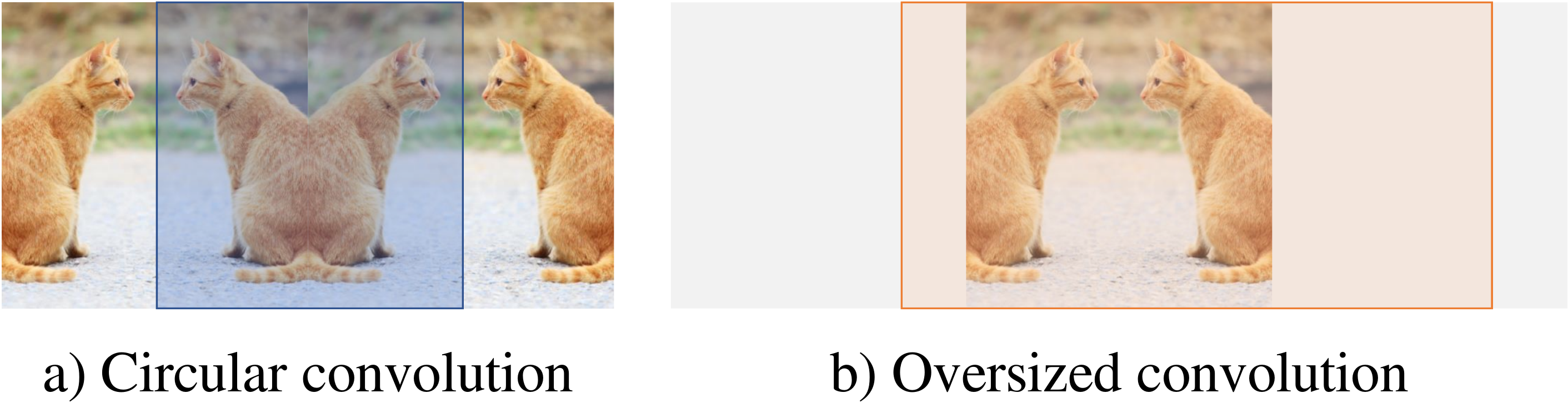}
  \caption{\textbf{Comparison between circular convolution and oversized convolution.} We only show horizontal convolution for illustration purposes.  a) Circular convolution in ParCNetV1 inevitably distorts context information at the boundary of images. b) Oversized convolution resolves the distortion while maintaining the global receptive field over the whole image.}
  \label{fig:circular_oversize}
  \vspace{-3mm}
\end{figure}

Despite improved model efficiency and accuracy, ParCNetV1 still suffers from some design drawbacks.
Firstly, as mentioned in~\cite{zhang2022parc} and shown in Fig~\ref{fig:circular_oversize}, the circular padding introduces spatial distortion by performing convolutions crossing image borders.
Secondly, the attention design is relatively weak compared with transformers which may limit the framework performance.
Thirdly, it is not feasible to apply global convolution to all blocks in CNNs, especially those shallow blocks due to expensive computational costs and over-smoothing effects.  

To address these issues, we propose a pure convolutional neural network architecture called ParCNetV2. It is composed of three essential improvements over ParCNetV1.

First, we push the kernel size to the extreme by doubling the circular convolution kernel and removing the absolute positional encoding. As shown in Fig.~\ref{fig:circular_oversize}, through large size (equal to the size of the input) padding, the convolution operation avoids feature distortion around image borders. By using constant paddings, the oversized kernel implicitly encodes spatial locations when it convolves with the feature maps~\cite{kayhan2020translation}. It enables us to discard the positional encoding module without hurting network performance. We explain why $2\times$ is the extreme in Sec.\ref{subsec:oversized}.

Second, the original ParC block uses a limited attention mechanism inserted at the end of the channel mixing phase.
We propose a more flexible bifurcate gate unit (BGU) 
at both the token mixing phase (spatial BGU) and channel mixing phase (channel BGU) in our newly designed block.
Compared to the squeeze-and-excitation block, the BGU is stronger while more compact and general to combine with various structures, leading to spatial attention and channel attention.
The enhanced attention mechanism also simplifies our ParC V2 block, as both phases adopt the consistent BGU structure.

Last, in contrast to ParCNetV1 which applies large kernel convolutions only on later-stage CNN blocks, we unify the block design by mixing large kernel convolutions with local depth-wise convolutions in all the blocks.
Both types of convolutions are operated on the input feature map channels.
This progressive design combines local features and global features in one convolution step, unlike many other works that stack the two sequentially~\cite{graham2021levit,xiao2021early,zhang2022parc} or as two separate branches~\cite{chen2022mobile,mehta2021mobilevit,dai2021coatnet}.
To this end, the resulting redesigned ParC V2 structure is capable of performing local convolutions, global convolutions, token channel mixing, and BGU-based attention all in one block.

To summarize, the main contributions of this paper are as follows:

\begin{itemize}
  \item We propose oversized convolutions for the effective modeling of long-range feature interactions in CNNs. Compared to ParCNetV1, it enables homogeneous convolution across all spatial locations, while removes the need for extra position encoding.

  \item We propose two bifurcate gate units (spatial BGU and channel BGU), which are compact and powerful attention modules. They boost the performance of ParCNetV2 and could be easily integrated into other network structures.

  \item We bring oversized convolution to shallow layers of CNNs and unify the local-global convolution design across blocks.
\end{itemize}

Extensive experiments are conducted to demonstrate that ParCNetV2 outperforms all other CNNs given a similar amount of parameters and computation budgets as shown in Fig.~\ref{fig:trade_offs}. It also beats state-of-the-art ViTs and CNN-ViT hybrids, which indicates that convolution networks are as strong as transformers in extracting features.

\section{Related Works}

\noindent\textbf{Convolution Networks.} Before transformers were introduced to vision tasks, convolutional neural networks had dominated vision architectures in a variety of computer vision tasks, such as image classification~\cite{krizhevsky2017imagenet,simonyan2014very,he2016deep}, object detection~\cite{ren2015faster,redmon2016you}, and semantic segmentation~\cite{chen2017deeplab,ronneberger2015u}. ResNet~\cite{he2016deep} introduced residual connections to eliminate network degradation, enabling very deep convolutional networks. It has been a strong baseline in various vision tasks. MobileNets~\cite{howard2017mobilenets,sandler2018mobilenetV2,howard2019searching} introduced depth separable convolution and ShuffleNets~\cite{zhang2018shufflenet,ma2018shufflenet} proposed group point-wise convolution with channel shuffling, both aimed to build light-weight models with small memory and computation footprint. After the appearance of vision transformers, researchers improved pure convolution networks with ideas from transformers. RepLKNet~\cite{ding2022scaling} increased kernel size to as large as $31\times 31$, which can extract long-range dependencies in contrast to commonly used $3\times 3$ kernels. ConvNeXt~\cite{liu2022convnet} reviewed the design of the vision transformers and gradually modernized a standard ResNet toward a transformer. They built a pure CNN model that competes favorably with the ViTs while maintaining the simplicity and efficiency of standard CNNs. ParCNet~\cite{zhang2022parc} proposed a pure convolution network with position-aware circular convolution, which achieved better performance than popular light-weight CNNs and vision transformers.

\noindent\textbf{Vision Transformers.} Dosovitskiy~\etal introduced the transformer model into vision tasks and proposed ViT~\cite{dosovitskiy2020image}. It cropped images into $16\times 16$ patches as input tokens to the transformer and used positional encoding to learn spatial information. However, the vanilla ViT was hard to train and huge datasets are required such as JFT-300M~\cite{sun2017revisiting}. DeiT~\cite{touvron2021training} exploited knowledge distillation to train ViT models and achieved competitive accuracy with less pretraining data. To further enhance the model architecture, some researchers attempted to optimize ViTs with ideas from CNNs. T2T-ViT~\cite{yuan2021tokens} introduced a token-to-token process to progressively tokenize images to tokens and structurally aggregate tokens. PVT~\cite{wang2021pyramid} inserted convolution into each stage of ViT to reduce the number of tokens and build hierarchical multi-stage structures. Swin transformer~\cite{liu2021swin} computed self-attention among shifted local windows, which has become the new baseline of many vision tasks. PiT~\cite{heo2021rethinking} jointly used pooling layers and depth-wise convolution layers to achieve channel multiplication and spatial reduction. Yu~\etal~\cite{yu2022metaformer} pointed out that the general architecture of the transformers is more essential to the model's performance instead of the specific token mixer module. They initiated the concept of MetaFormer which is compatible with using convolutions, self-attention, and even pooling as the token mixer.

\noindent\textbf{Hybrid Convolution Networks and Vision Transformers.} In addition to ViTs, another popular line of research is to combine elements of ViTs and CNNs to absorb the strengths of both architectures. LeViT~\cite{graham2021levit} proposed a hybrid neural network for fast inference and significantly outperformed existing CNNs and ViTs concerning the speed/accuracy trade-off. BoTNet~\cite{srinivas2021bottleneck} replaces the standard convolutions with multi-head attention in the final three bottleneck blocks of ResNet. CvT~\cite{wu2021cvt} introduced depth-wise and point-wise convolution in front of the self-attention unit, which introduced shift, scale, and distortion invariance while maintaining the merits of transformers. Some other works focused on improving efficiency with hybrid models. CMT~\cite{guo2022cmt} combined a convolutional inverted feed-forward network with a lightweight multi-head self-attention way and took advantage of transformers to capture long-range dependencies and CNN to model local features. MobileViT~\cite{mehta2021mobilevit} proposed a lightweight model and a fast training strategy for mobile devices. Mobile-Former~\cite{chen2022mobile} adopted a parallel structure to combine convolution blocks and attention blocks.

Although many works have successfully combined transformers and CNNs for vision tasks, they are not as much focused as our work on the systematic design of the global receptive field, advanced attention mechanism, and unified local-global balance across the whole network. We invent a newly evolved version of these designs and demonstrate the potential of pure CNNs compared with transformers and hybrid architectures.

\begin{figure*}
  \centering
  \includegraphics[width=0.98\linewidth]{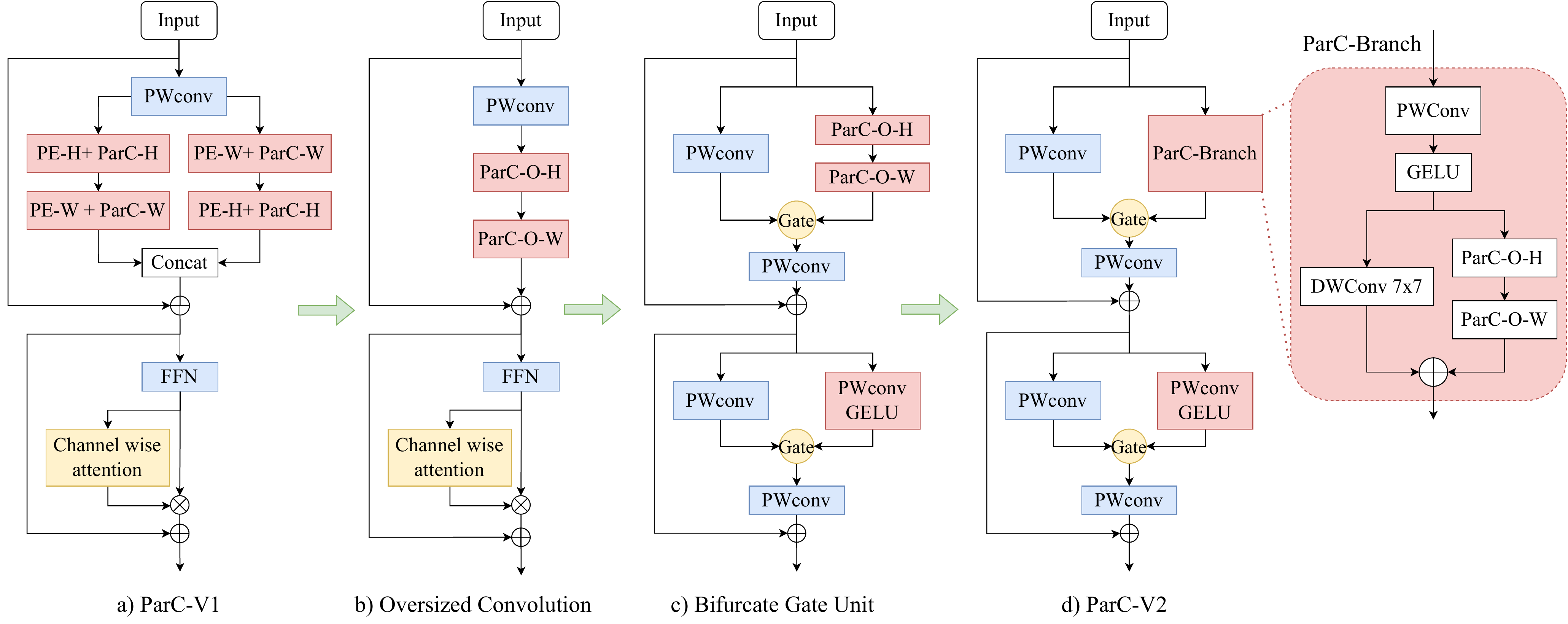}
  \caption{\textbf{The transitions from the original ParC V1 to ParC V2 block.} Compared with ParCNetV1, we first introduce oversized convolutions to further enhance capacity while simplify architecture; then we design bifurcate gate unit to improve efficiency and strengthen attention; finally we propose uniform local-global block and construct the whole network with this uniform block.}
  \label{fig:evolution}
  \vspace{-3mm}
\end{figure*}

\section{Methods}

An overview of the ParCNetV2 architecture is presented in Fig.~\ref{fig:evolution}.
Compared with the original ParCNet (Fig.~\ref{fig:evolution}a), we first substitute the position-aware circular convolution with oversized convolution to encode long-range dependencies along with position information (Fig.~\ref{fig:evolution}b).
Then we introduce bifurcate gate units as a stronger attention mechanism (Fig.~\ref{fig:evolution}c).
Finally, we propose a uniform block that balances local and global convolutions to build full ParCNetV2 (Fig.~\ref{fig:evolution}d).
The following sections describe the details of these components.

\subsection{Oversized convolution}
\label{subsec:oversized}

In ParCNetV1, the model is divided into two branches, alternating the order of vertical and horizontal convolution.
However, we find that changing the order does not affect the output (proof in supplementary), thus we keep only one branch for simplicity.
To further enhance the model's capacity and incorporate long-range spatial context, we introduce an oversized depth-wise convolution with a kernel size approximately twice the input feature size (ParC-O-H and ParC-O-W), as illustrated in Fig.~\ref{fig:evolution}b.
In this section, we provide details about the oversized convolution and discuss its effectiveness, efficiency, and adaptability.

\begin{figure}
  \centering
  \includegraphics[width=0.9\linewidth]{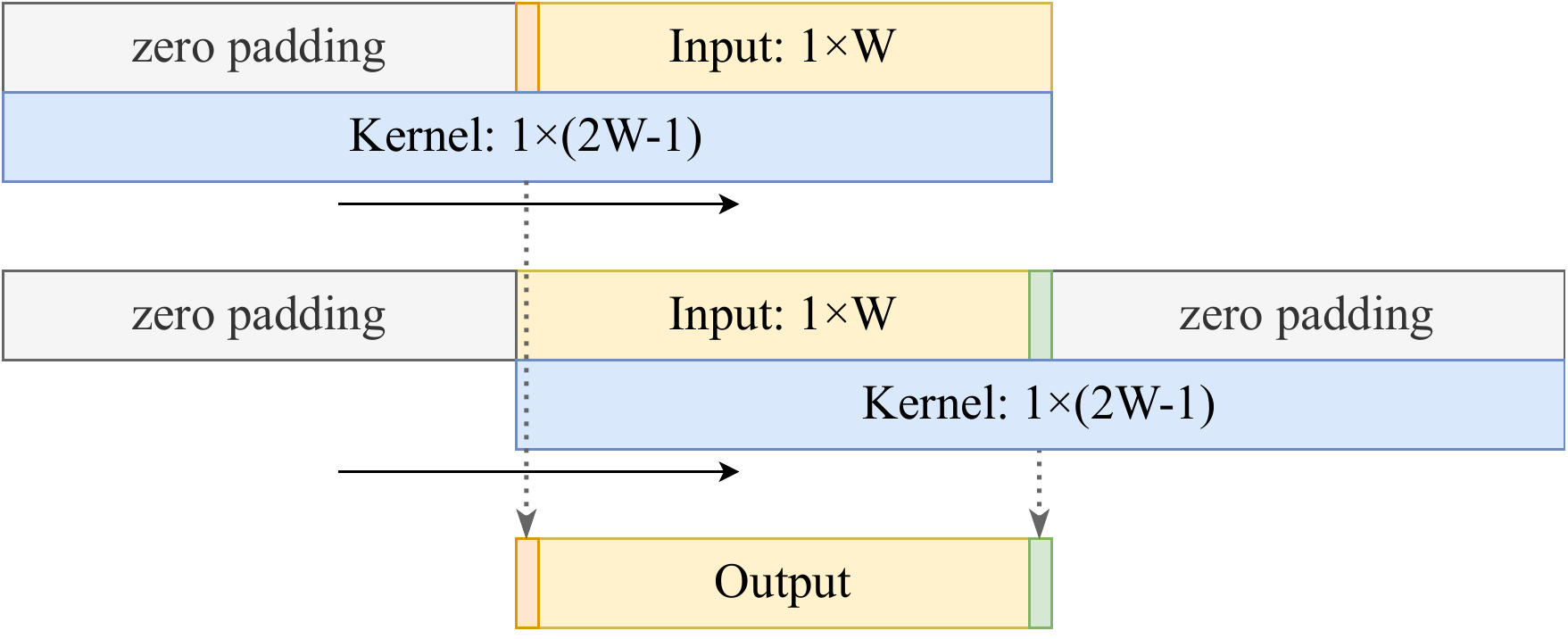}
  \caption{\textbf{Illustration of the oversized convolution.} Kernels are almost twice the size of input feature maps, and zero-padding is applied to keep the output resolution the same as the input.}
  \label{fig:oversize_conv}
  \vspace{-3mm}
\end{figure}

\noindent\textbf{Formulation:}
We denote the input feature map as $X\in \mathcal{R}^{C\times H \times W}$, where $C$, $H$, and $W$ represent the number of channels, height, and width of $X$, respectively.
The kernel weight for vertical and horizontal oversized convolution is $k^{h}\in\mathcal{R}^{C\times (2H - 1)\times 1}$ and $k^{w}\in\mathcal{R}^{C\times 1\times (2W - 1)}$.
We let index $0$ denote the center point of $k^{h}$ and $k^{w}$.
As shown in Fig.~\ref{fig:oversize_conv}, we choose this size because it naturally covers the global receptive field at each position, and keeps the output size the same as the input without requiring any post-processing. In contrast, smaller kernels can not simultaneously preserve position cues and provide a global receptive field, while larger kernels need post-processing to adjust the output size.

To compute the output of the oversized convolution $Z_{i,j}$ at location $(i, j)$, we use the following equations:
\begin{align}
  Y_{i,j} & = \sum_{s=-(H - 1)}^{H - 1}k^h_s X_{i+s, j},\label{eq:parc_h} \\
  Z_{i,j} & = \sum_{t=-(W - 1)}^{W - 1}k^w_t Y_{i, j+t},\label{eq:parc_w}
\end{align}
where Eq.~\eqref{eq:parc_h} denotes ParC-O-H, and Eq.~\eqref{eq:parc_w} denotes ParC-O-W.
Zero-padding means that $X_{i, j} = 0$ and $Y_{i, j} = 0$, if $i\notin [0, H - 1]$ or $j\notin[0, W - 1]$.

The padding operation is designed to work with oversized convolution, which encodes not only global dependency but also position information.
For the horizontal convolution, we apply $W-1$ pixels zero padding to both left and right sides, where $W$ is the width of the input feature.
Similar operations are performed for vertical convolution.
This schema keeps the output feature size the same as the input feature, and implicitly encodes position cues by zeroing out partial convolution kernel parameters according to spatial locations.

\noindent\textbf{Effectiveness:}
The oversized convolution brings two advantages.
First, it encodes position information by embedding it into each location using zero-padding, eliminating the need for position embeddings.
As shown in Fig.~\ref{fig:oversize_conv}, each position in the output is transformed by different parameters across the input features, and thus embeds position information in the model weights.
It is similar to relative position embeddings~\cite{shaw2018self}, while the oversized convolution encodes both spatial context and position information in kernel weights.
As a result, position embeddings are no longer required and therefore abandoned to make the network more concise.

Second, it improves model capacity with limited computational complexity.
For instance, the largest oversized kernel in ParCNetV2-Tiny is extended to $111\times 1 $ and $1\times 111$ with input size $224\times 224$.
The capacity of the model will be significantly enhanced with such large convolution kernels.
As far as we know, it has achieved the largest convolution kernel among prevailing vision CNNs.
Other works on large kernel~\cite{rao2021global,ding2022scaling,liu2022more} use a spatially dense form of convolution, which requires massive computation.
In contrast, our oversized convolution boosts performance with much less computation cost.
It enables our model to achieve state-of-the-art performance, which indicates that it is an effective operation.

\noindent\textbf{Efficiency:}
Although the oversized convolution has less computation than the previous large kernel convolution networks~\cite{ding2022scaling,liu2022more}, the multi-fragment structure is poorly supported by the hardware, especially with PyTorch. This is because PyTorch is not optimized for multi-fragmentation, hence we implement a block-wise (inverse) \textit{implicit gemm} algorithm following RepLKNet~\cite{ding2022scaling}. Fig~\ref{fig:trade_offs} shows the comparison results. Compared to other recently proposed models, our ParCNetV2 offers a clear advantage in terms of both accuracy and inference speed. Furthermore, \textit{even on Vanilla PyTorch, our ParCNetV2 achieves a superior trade-off between accuracy and speed}. Additional results can be found in the supplementary material.


\noindent\textbf{Adaptability to multi-scale input:}
To deal with input images of different resolutions, each convolution kernel will be first zoomed with linear interpolation to $C\times (2H - 1)\times 1$ and $C\times 1 \times (2W - 1)$.
In addition, this method keeps the model's global receptive field on any input size and learns to extract scale-invariant features.

\subsection{Bifurcate Gate Unit}

\begin{figure}
  \centering
  \includegraphics[width=0.9\linewidth]{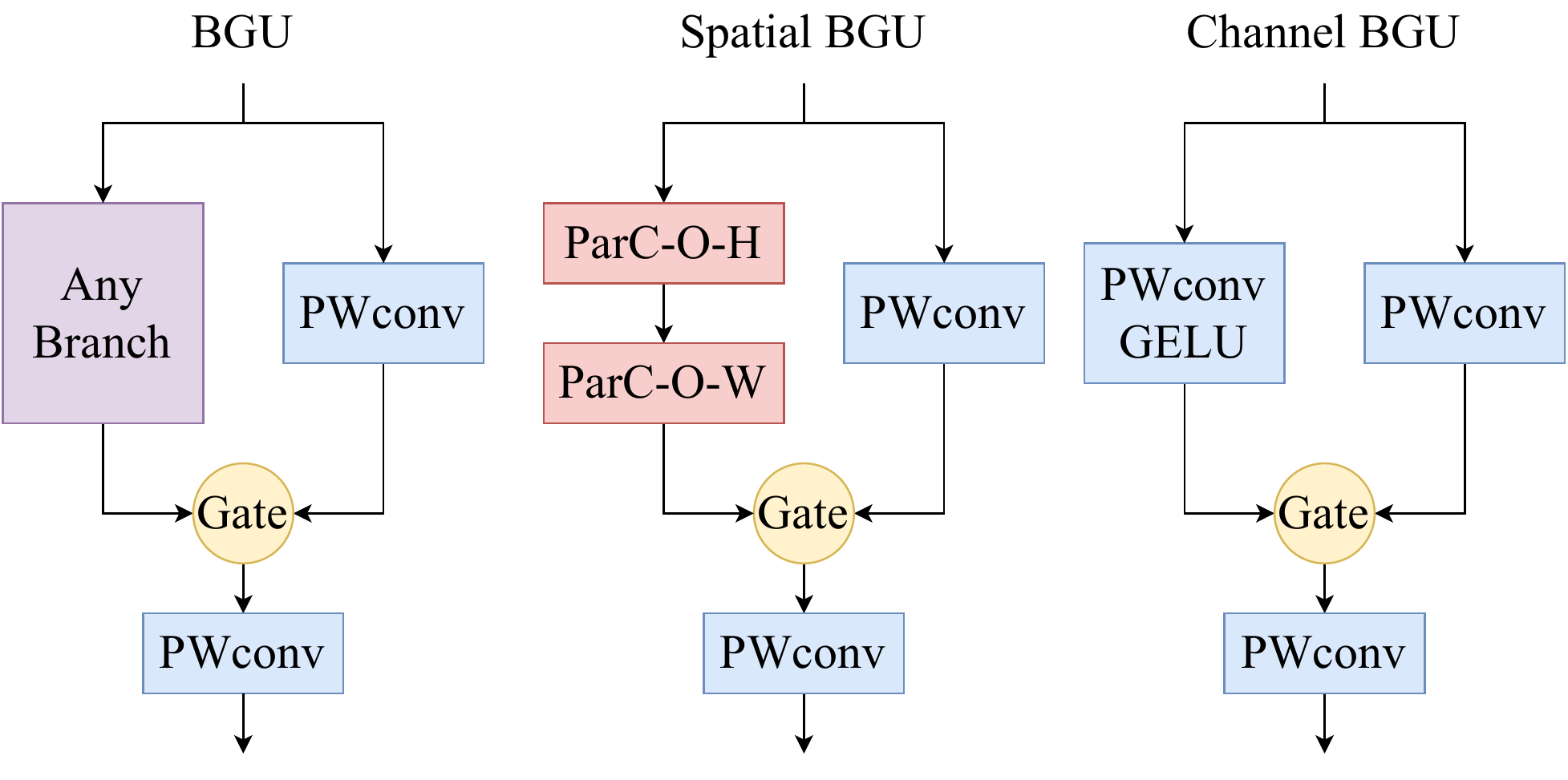}
  \caption{\textbf{Illustration of the Bifurcate gate unit (BGU).} We propose a general BGU which can be easily integrated into various network structures. For Spatial GPU, we insert our ParC branch and a point-wise convolution to extract spatial features. While in Channel BGU, we simply adopt a point-wise convolution to conduct channel mixing.}
  \label{fig:bgu}
  \vspace{-3mm}
\end{figure}

To make the model data-driven as ViT models, ParCNetV1 employed the squeeze-and-excitation block, which was demonstrated to boost the model performance on various tasks.
In this work, the attention mechanism is reinvented with two major improvements: strengthened attention and better computation efficiency.
Specifically, we propose the Bifurcate Gate Unit (BGU) structure inspired by gated linear unit (GLU)~\cite{dauphin2017language} which improves MLP through gating mechanism.
BGU inherits high computation efficiency from GLU and accomplishes attention and feature extraction in a single unit.
Different from GLU which inserts gate operation into two homologous features, the proposed BGU applies gate operation on two features from two branches. 
One branch adopts a point-wise convolution to serve as attention weights.
The other transforms the features depending on the purpose of the module, \emph{i.e.}, ParC branch to extract spatial information for spatial interaction, and point-wise convolution to perform channel mixing.
Therefore, the BGU design is extended to spatial BGU and channel BGU modules, making it a general module as shown in Fig.~\ref{fig:bgu}.
Finally, the outputs of the two branches are fused by an element-wise multiplication operation and an additional point-wise convolution.
We introduce the details and discuss the difference from other attentions in this section.

\noindent\textbf{Spatial BGU:}
In the spatial BGU, we aim to extract representative spatial information including local and global dependencies. We adopt ParC branch as the feature transform branch, which consists of a point-wise convolution, a standard local depth-wise convolution and an oversized separable convolution. We will describe it in detail in Sec.~\ref{subsec:uniform}. Basically, our spatial BGU is defined as:
\begin{align*}
   & X_1 = \operatorname{ParC}(X),                                         \\
   & X_2 = \operatorname{PWConv_1}(X),                                     \\
   & \operatorname{SpatialBGU}(X) = \operatorname{PWConv_2}(X_1\odot X_2).
\end{align*}

\noindent\textbf{Channel BGU:}
For the channel mixing module, the original feed-forward network (FFN) of common transformers usually contains two point-wise convolutions separated by a GELU activation. The first layer expands the number of channels by a factor of $\alpha$, and the second layer shrinks the dimension back to the original:
\begin{equation*}
  \operatorname{FFN}(X) = \operatorname{GELU}(XW_1+b_1)W_2+b_2,
\end{equation*}
where $W_1\in\mathbf{R}^{C\times \alpha C}$ and $W_2\in\mathbf{R}^{\alpha C\times C}$ indicate weights of the two point-wise convolutions, $b_1$ and $b_2$ are the bias terms, respectively.
In our channel BGU, we split the hidden layer into two branches and merge with element-wise multiplication. The whole module is defined as:
\begin{align*}
   & X_1 = \operatorname{GELU}(X\widetilde{W}_1+\Tilde{b}_1),                    \\
   & X_2 = X\widetilde{W}_2+\Tilde{b}_2,                                         \\
   & \operatorname{ChannelBGU}(X) = (X_1\odot X_2)\widetilde{W}_3 + \Tilde{b}_3,
\end{align*}
where $\widetilde{W}_1, \widetilde{W}_2\in\mathbf{R}^{C\times \Tilde{\alpha}C}$ and $\widetilde{W}_3\in\mathbf{R}^{\Tilde{\alpha}C\times C}$ indicates weights of point-wise convolutions, $\Tilde{b}_1$, $\Tilde{b}_2$, $\Tilde{b}_3$ denotes biases, respectively. We adjust $\Tilde{\alpha}$ to fit the model size close to the original FFN (details in supplementary).

\noindent\textbf{Comparisons with previous attention mechanisms:}
The classic channel attentions~\cite{hu2018squeeze,wang2020eca,qin2021fcanet} and spatial attentions~\cite{woo2018cbam,hou2021coordinate} consist of two imbalanced branches: a heavy backbone branch and a light attention branch.
The attention branch drops massive information by global average pooling, shared attention value across channels or space, and bottleneck structures.
However, it contains a large number of parameters similar to the backbone branch. BGU is a compact attention mechanism with more balanced branches.
There is no downsampling or bottleneck in each branch. Besides, BGU does not increase the number of parameters of the model.

\subsection{Uniform local-global convolution}
\label{subsec:uniform}
ParCNetV1 used two different network structures, traditional convolutional block MBConvs~\cite{howard2019searching} in shallow layers and ParC operation in deep layers.
We extend the global convolution to each block through the early and late stage, since it is shown that a large receptive field is also critical in the shallow layers, especially in downstream tasks~\cite{ding2022scaling,liu2022more}.
We design a unified block composed of both local and global convolutions for the entire network.
As shown in Fig.~\ref{fig:evolution}, we adopt a point-wise convolution first to fuse channel information.
Then we pass the feature into two branches, one of which is a standard 7x7 depth-wise convolution to extract local cues, and the other is an oversized convolution to model global independence.
Finally, we add the two branches to create a multiscale feature.
Formally, the uniform local-global convolution is defined as:
\begin{align*}
  Y_{local}              & = \operatorname{DWConv}\left(X\right),
  \\
  Y_{global}             & = \operatorname{ParC-O-W}\left(\operatorname{ParC-O-H}\left(X\right)\right), \\
  \operatorname{ParC}(X) & = \operatorname{PWConv}\left(Y_{local} + Y_{global}\right).
\end{align*}

\subsection{ParCNetV2}
\label{sec::fullnet}

\begin{table}
  \begin{center}
    \setlength{\tabcolsep}{10pt}
    \small
    \begin{tabular}{l|ccccc}
      \toprule
      Models       & No. Channels        & No. Blocks    \\ 
      \hline
      ParCNetV2-XT & (48, 96, 192, 320)  & (3, 3, 9, 2)  \\ 
      ParCNetV2-T  & (64, 128, 320, 512) & (3, 3, 12, 3) \\ 
      ParCNetV2-S  & (64, 128, 320, 512) & (3, 9, 24, 3) \\ 
      ParCNetV2-B  & (96, 192, 384, 576) & (3, 9, 24, 3) \\ 
      \bottomrule
    \end{tabular}
  \end{center}
  \caption{\textbf{Model configuration of ParCNetV2.} Each tuple represents the number of channels or blocks for the four stages.}
  \label{tab:configuration}
  \vspace{-3mm}
\end{table}

Based on the proposed modules above, we build ParCNetV2 with four different scales. We adopt a hierarchical architecture with 4-stage inspired by~\cite{liu2021swin,liu2022convnet}, and the number of channels and blocks of each stage are listed in Tab.~\ref{tab:configuration}. ParCNetV2-XT is designed to fairly compare with ParC-ConvNeXt-T ($0.5\times$W), which is a four-stage version of ParCNetV1~\cite{zhang2022parc}.
ParCNetV2-T, ParCNetV2-S, and ParCNetV2-B are designed to compare with the state-of-the-art networks.
The expand ratio $\Tilde{\alpha}$ of channel BGU is set to 2.5, which is close to the original FFN in complexity.

\section{Experiments}

In this section, we exhibit quantitative and qualitative experiments to demonstrate the effectiveness of the proposed model. First of all, we conduct experiments on image classification on the ImageNet-1K~\cite{deng2009imagenet}. We compare the performance with convolutional neural networks and show that our ParCNetV2 performs better over pure convolutional networks, including ParCNetV1. Then, we compare our model with transformers and hybrid neural networks. Next, we conduct experiments on downstream tasks including object detection and instance segmentation on COCO~\cite{lin2014microsoft}, and semantic segmentation on ADE20K dataset~\cite{zhou2019semantic}. Finally, we compare the inference latency on GPUs and edge devices. All experiments are implemented based on PyTorch~\cite{paszke2019pytorch}.

\begin{table}
  \renewcommand\arraystretch{1.0}
  \begin{center}
    \setlength{\tabcolsep}{1pt}
    \small
    \begin{tabular}{l|ccc}
      \toprule
      Models                                                         & Param(M) & MACs(G) & Top-1(\%)     \\
      \hline
      ReGNetY-1.6G~\cite{radosavovic2020designing}                   & 11       & 1.6     & 78.0          \\
      ParC-Net-S                    ~\cite{zhang2022parc}            & 5.0      & 3.5     & 78.6          \\ %
      ParC-ConvNeXt-T($0.5\times$W)~\cite{zhang2022parc}             & 7.4      & 1.1     & 78.3          \\ %
      \rowcolor{cyan!10}
      ParCNetV2-XT                                                   & 7.4      & 1.6     & \textbf{79.4} \\ %
      \hline
      ResNet50~\cite{he2016deep,wightman2021resnet}                  & 23       & 4.1     & 79.8          \\
      ReGNetY-4G~\cite{radosavovic2020designing,wightman2021resnet}  & 21       & 4.0     & 81.3          \\
      ResNeSt50~\cite{zhang2022resnest}                              & 28       & 5.4     & 81.1          \\
      ConvNeXt-T~\cite{liu2022convnet}                               & 29       & 4.5     & 82.1          \\
      SLaK-T~\cite{liu2022more}                                      & 30       & 5.0     & 82.5          \\
      PoolFormer-S24~\cite{yu2022metaformer}                         & 21       & 3.6     & 80.3          \\
      ParCNetV1-27M~\cite{zhang2022parc}                             & 27       & 4.5     & 82.1          \\
      \rowcolor{cyan!10}
      ParCNetV2-T                                                    & 25       & 4.3     & \textbf{83.5} \\
      \hline
      ResNet101~\cite{he2016deep,wightman2021resnet}                 & 45       & 7.9     & 81.3          \\
      ReGNetY-8G~\cite{radosavovic2020designing,wightman2021resnet}  & 39       & 8.0     & 82.1          \\
      ConvNeXt-S~\cite{liu2022convnet}                               & 50       & 8.7     & 83.1          \\
      SLaK-S~\cite{liu2022more}                                      & 55       & 9.8     & 83.8          \\
      \rowcolor{cyan!10}
      ParCNetV2-S                                                    & 39       & 7.8     & \textbf{84.3} \\
      \hline
      ResNet152~\cite{he2016deep,wightman2021resnet}                 & 60       & 11.6    & 81.8          \\
      ReGNetY-16G~\cite{radosavovic2020designing,wightman2021resnet} & 84       & 15.9    & 82.2          \\
      ConvNeXt-B~\cite{liu2022convnet}                               & 89       & 15.4    & 83.8          \\
      RepLKNet-31B~\cite{ding2022scaling}                            & 79       & 15.3    & 83.5          \\
      SLaK-B~\cite{liu2022more}                                      & 95       & 17.1    & 84.0          \\
      \rowcolor{cyan!10}
      ParCNetV2-B                                                    & 56       & 12.5    & \textbf{84.6} \\
      \bottomrule
    \end{tabular}
  \end{center}
  \caption{\textbf{Comparison with the modern convolution networks on image classification.} All experiments are trained on ImageNet-1K dataset with 300 epochs. Top-1 accuracy on the validation set is reported. \textbf{ParC-ConvNeXt-T ($0.5\times$W)}~\cite{zhang2022parc}: ParCNetV1 of hierarchical 4-stage architecture the same as ParCNetV2. \textbf{ParCNetV1-27M}: ParCNetV1 with bigger backbone.}
  \label{tab:pure_conv}
  \vspace{-3mm}
\end{table}

\subsection{Performance Comparison with CNNs}
\label{subsec:pure_conv}

We conduct image classification on ImageNet-1K~\cite{deng2009imagenet}, the most widely used benchmark dataset. We train the ParCNetV2 models on the training set and report top-1 accuracy on the validation set.
We follow the same training hyperparameters and augmentations used in ConvNeXt~\cite{liu2022convnet} except that the batch size is restricted to 2048 and the initial learning rate is set to $4\times-3$. We also substitute LayerScale with Resscale~\cite{shleifer2021normformer} to stabilize training.

The comparison with pure convolution networks on image classification is listed in Tab.~\ref{tab:pure_conv}.
It is clear that ParCNetV2 outperforms other convolutional networks by a large margin across various model scales, including variants of the ResNet (ResNet~\cite{he2016deep,wightman2021resnet}, 
ResNeSt~\cite{zhang2022resnest}), NAS architecture (ReGNetY~\cite{radosavovic2020designing}), ConvNeXt~\cite{liu2022convnet}, and MetaFormer architecture (PoolFormer~\cite{yu2022metaformer}).
Specifically, our ParCNetV2-T surpasses ParCNetV1-27M~\cite{zhang2022parc}, which indicates that our methods go deeper along the larger convolutions and stronger attention mechanisms.
In addition, ParCNetV2-S performs better than all the other CNNs even twice larger in parameters and complexity, which indicates our model is highly effective.

\begin{table}
  \renewcommand\arraystretch{0.9}
  \begin{center}
    \setlength{\tabcolsep}{5.6pt}
    \small
    \begin{tabular}{l|c|cccc}
      \toprule
      Models                               & Mixing      & Param & MACs & Top-1         \\
                                           & Type        & (M)   & (G)  & (\%)          \\
      \hline
      DeIT-S~\cite{touvron2021training}    & Attn        & 22    & 4.6  & 79.9          \\
      T2T-ViT-14~\cite{yuan2021tokens}     & Attn        & 21.5  & 4.8  & 81.5          \\
      Swin-T~\cite{liu2021swin}            & Attn        & 29    & 4.5  & 81.3          \\
      CSwin-T~\cite{dong2022cswin}         & Attn        & 23    & 4.3  & 82.7          \\
      CvT-13~\cite{wu2021cvt}              & Attn + Conv & 20    & 4.5  & 81.6          \\
      CoAtNet-0~\cite{dai2021coatnet}      & Attn + Conv & 25    & 4.2  & 81.6          \\
      Next-ViT-S~\cite{li2022next}         & Attn + Conv & 32    & 5.8  & 82.5          \\
      Uniformer-S~\cite{li2022uniformer}   & Attn + Conv & 20    & 4.8  & 82.9          \\
      \rowcolor{cyan!10}
      ParCNetV2-T                          & Conv        & 25    & 4.3  & \textbf{83.5} \\
      \hline
      T2T-ViT-19~\cite{yuan2021tokens}     & Attn        & 39    & 8.5  & 81.9          \\
      Swin-S~\cite{liu2021swin}            & Attn        & 50    & 8.7  & 83.0          \\
      CSwin-S~\cite{dong2022cswin}         & Attn        & 35    & 6.9  & 83.6          \\
      CvT-21~\cite{wu2021cvt}              & Attn + Conv & 32    & 7.1  & 82.5          \\
      CoAtNet-1~\cite{dai2021coatnet}      & Attn + Conv & 42    & 8.4  & 83.3          \\
      Next-ViT-B~\cite{li2022next}         & Attn + Conv & 45    & 8.3  & 83.2          \\
      Uniformer-B~\cite{li2022uniformer}   & Attn + Conv & 50    & 8.3  & 83.9          \\
      \rowcolor{cyan!10}
      ParCNetV2-S                          & Conv        & 39    & 7.8  & \textbf{84.3} \\
      \hline
      DeiT-B/16~\cite{touvron2021training} & Attn        & 86    & 17.6 & 81.8          \\
      T2T-ViT-24~\cite{yuan2021tokens}     & Attn        & 64    & 13.8 & 82.3          \\
      Swin-B~\cite{liu2021swin}            & Attn        & 88    & 15.4 & 83.5          \\
      CSwin-B~\cite{dong2022cswin}         & Attn        & 78    & 15.0 & 84.2          \\
      CoAtNet-2~\cite{dai2021coatnet}      & Attn + Conv & 75    & 15.7 & 84.1          \\
      Next-ViT-L~\cite{li2022next}         & Attn + Conv & 58    & 10.8 & 83.6          \\
      \rowcolor{cyan!10}
      ParCNetV2-B                          & Conv        & 56    & 12.5 & \textbf{84.6} \\
      \bottomrule
    \end{tabular}
  \end{center}
  \caption{\textbf{Comparison with state of the art transformer and hybrid networks on ImageNet-1K classification dataset.} Top-1 accuracy on the validation set is reported.}
  \label{tab:classification}
  \vspace{-3mm}
\end{table}

\subsection{Performance Comparison with ViTs and Hybrid Models}
\label{subsec:classification}

Apart from CNNs, ParCNetV2 also beats various latest ViTs and Hybrid models. As shown in Tab.~\ref{tab:classification}, compared with famous transformers such as Swin-T~\cite{liu2021swin} and CSwin-T~\cite{dong2022cswin}, ParCNetV2-T improves the accuracy by a clear margin of 2.2\% and 0.8\% with comparable parameters and computational costs. This result demonstrates that our pure convolution model utilizes the design concepts from transformers in a more efficient way. Compared with hybrid models, ParCNetV2-T outperforms CvT~\cite{wu2021cvt}, CoAtNet~\cite{dai2021coatnet}, Uniformer~\cite{li2022uniformer} and Next-ViT~\cite{li2022next} with much fewer parameters. Combined with the above analysis of pure convolutions in Sec.~\ref{subsec:pure_conv}, our proposed model has achieved better classification accuracy with comparable parameters and computation sizes over various kinds of architectures.

\subsection{ParC V2 Performance on Downstream Tasks}

To evaluate the transfer ability of ParC V2, we conduct experiments on the object detection and instance segmentation task with COCO~\cite{lin2014microsoft} semantic segmentation task with ADE20K~\cite{zhou2019semantic}.

\begin{table}
  \begin{center}
  \renewcommand\arraystretch{0.95}
  \setlength{\tabcolsep}{0.75pt}
  \small
  \begin{tabular}{l|cccccc}
    \toprule
    backbone                         & AP$^{\mathit{bbox}}$                                      & AP$^{\mathit{bbox}}_{50}$ & AP$^{\mathit{bbox}}_{75}$ & AP$^{\mathit{mask}}$ & AP$^{\mathit{mask}}_{50}$ & AP$^{\mathit{mask}}_{75}$ \\
    \hline
                                     & \multicolumn{6}{c}{Mask R-CNN $3\times$ schedule}                                                                                                                                                \\
    Swin-T~\cite{liu2021swin}        & 46.0                                                      & 68.1                      & 50.3                      & 41.6                 & 65.1                      & 44.9                      \\
    ConvNeXt-T~\cite{liu2022convnet} & 46.2                                                      & 67.9                      & 50.8                      & 41.7                 & 65.0                      & 44.9                      \\
    \rowcolor{cyan!10}
    ParCNetV2-T                      & \textbf{48.9}                                             & 70.3                      & 53.9                      & \textbf{43.7}        & 67.6                      & 47.0                      \\
    \hline
                                     & \multicolumn{6}{c}{Cascade Mask R-CNN $3\times$ schedule}                                                                                                                                        \\
    Swin-T~\cite{liu2021swin}        & 50.4                                                      & 69.2                      & 54.7                      & 43.7                 & 66.6                      & 47.3                      \\
    ConvNeXt-T~\cite{liu2022convnet} & 50.4                                                      & 69.1                      & 54.8                      & 43.7                 & 66.5                      & 47.3                      \\
    \rowcolor{cyan!10}
    ParCNetV2-T                      & \textbf{52.6}                                             & 71.0                      & 57.3                      & \textbf{45.6}        & 68.6                      & 49.8                      \\
    \bottomrule
  \end{tabular}  
  \end{center}
  \caption{Comparisons on \textbf{COCO~\cite{lin2014microsoft} object detection and instance segmentation.} We use Mask R-CNN and Cascade Mask R-CNN~\cite{cai2018cascade} as a basic framework. All models are pretrained on ImageNet-1K and trained on COCO for $3\times$ iterations.}
  \label{tab:detection}
  \vspace{-0mm}
\end{table}

\begin{table}
  \begin{center}
    \renewcommand\arraystretch{0.95}
    \setlength{\tabcolsep}{6pt}
    \small
    \begin{tabular}{l|cccc}
      \toprule
      backbone                                & Param(M) & MACs(G) & mIoU(\%)      \\
      \hline
      Swin-T~\cite{liu2021swin}               & 60       & 945     & 45.8          \\
      ConvNeXt-T~\cite{liu2022convnet}        & 60       & 939     & 46.7          \\
      ParCNetV1-27M~\cite{zhang2022parc}      & 56       & 936     & 46.7          \\
      Deit III (ViT-S)~\cite{touvron2022deit} & 42       & 588     & 46.8          \\
      \rowcolor{cyan!10}
      ParCNetV2-T                             & 55       & 932     & \textbf{49.4} \\  
      \hline
      Swin-S~\cite{liu2021swin}               & 81       & 1038    & 49.5          \\
      ConvNeXt-S~\cite{liu2022convnet}        & 82       & 1027    & 49.6          \\
      Deit III (ViT-B)~\cite{touvron2022deit} & 128      & 1283    & 50.2          \\
      \rowcolor{cyan!10}
      ParCNetV2-S                             & 69       & 1005    & \textbf{51.0} \\  
      \bottomrule
    \end{tabular}
  \end{center}
  \caption{Comparisons on \textbf{ADE20K~\cite{zhou2019semantic} semantic segmentation.} We use UperNet as a basic framework. All models are pretrained on ImageNet-1K and trained on ADE20K for 160K iterations. MACs are measured with the input size of (2048, 512).} 
  \label{tab:segmentation}
  \vspace{-3mm}
\end{table}

\noindent\textbf{Object detection and instance segmentaion on COCO.}
Following previous works~\cite{liu2021swin,liu2022convnet}, we finetune Mask R-CNN and Cascade Mask R-CNN~\cite{cai2018cascade} on COCO dataset~\cite{lin2014microsoft} with ParCNetV2 backbones. MMDetection~\cite{chen2019mmdetection} is used as the base framework. All models use pre-trained weights from ImageNet1K and are trained with $3\times$ schedule with multi-scale training. The experiment settings follow~\cite{liu2022convnet}.
Tab.~\ref{tab:detection} shows object detection and instance segmentation results comparing our ParCNetV2 with Swin~\cite{liu2021swin} and ConvNeXt~\cite{liu2022convnet}. ParCNetV2 outperforms both the transformer network and convolution network by a large margin across different model complexities. Interestingly, in experiments using Cascade Mask R-CNN, ParCNetV2-T has already outperformed larger models such as Swin-S and ConvNeXt-S, achieving 51.9 AP$^{\mathit{bbox}}$ and 45.0 AP$^{\mathit{mask}}$, which is a significant improvement of +0.7 AP$^{\mathit{bbox}}$ and +0.6 AP$^{\mathit{mask}}$, respectively. For further information on experiments with backbones of different scales, please refer to the supplementary materials.


\noindent\textbf{Semantic segmentation on ADE20K.} We finetune UperNet~\cite{xiao2018unified} on the ADE20K~\cite{zhou2019semantic} dataset with ParCNetV2 backbones.
MMSegmentation~\cite{contributors2020mmsegmentation} is used as the base framework.
All models use pre-trained weights from ImageNet1K and are trained for 160K iterations with a batch size of 16.
Experiment settings follow~\cite{liu2022convnet}.
Tab.~\ref{tab:segmentation} lists the mIoU, model size, and MACs for different backbones. ParCNetV2 achieves a substantially higher mIoU than Swin and ConvNeXt, while taking fewer parameters and computation.
Specifically, our model is +2.7\% mIoU higher than ParCNetV1-27M~\cite{zhang2022parc}, which validates the transferability of our ParCNetV2 model.

\subsection{Ablation Study}

\begin{table}
  \renewcommand\arraystretch{0.9}
  \begin{center}
    \small
    \setlength{\tabcolsep}{2pt}
    \begin{tabular}{c|cccc|ccc}
      \toprule
      \multirow{2}{*}{Row} & \multirow{2}{*}{OC} & \multirow{2}{*}{S-BGU} & \multirow{2}{*}{C-BGU} & \multirow{2}{*}{Uniform} & Param & MACs & Top-1         \\ %
                           &                     &                        &                        &                          & (M)   & (G)  & (\%)          \\ %
      \hline
      \rowcolor{cyan!10}
      baseline             & \checkmark          & \checkmark             & \checkmark             & \checkmark               & 7.4   & 1.6  & \textbf{79.4} \\ %
      \hline
      1                    &                     & \checkmark             & \checkmark             & \checkmark               & 7.2   & 1.4  & 78.9          \\ %
      2                    & \checkmark          &                        & \checkmark             & \checkmark               & 7.4   & 1.6  & 79.2          \\ %
      3                    & \checkmark          & \checkmark             &                        & \checkmark               & 7.4   & 1.5  & 79.1          \\ %
      4                    & \checkmark          & \checkmark             & \checkmark             &                          & 7.4   & 1.4  & 79.2          \\ %
      \bottomrule
    \end{tabular}
  \end{center}
  \caption{\textbf{Ablation study of each component on the ImageNet-1K classification task.} We use smaller ParCNetV2-XT in ablation for fast evaluation. Top-1 accuracy on the validation set is reported.
    \textbf{OC}: Oversized Convolution.
    \textbf{S-BGU}: Spatial Bifurcate Gate Unit.
    \textbf{C-BGU}: Channel Bifurcate Gate Unit.
    \textbf{Uniform}: Uniform local-global convolution. }
  \label{tab:ablation_components}
  \vspace{-3mm}
\end{table}

In this section, we make an ablation study on ImageNet-1K classification to show that each component in our ParCNetV2 is critical. To speed up the experiment, we use the smaller ParCNetV2-XT in this section. Training settings are the same as image classification experiments in Sec.~\ref{subsec:classification}.


\noindent\textbf{Oversized convolution.} Oversized convolution increases the capacity of the model and encodes position information. Without oversized convolution, the model not only loses capacity and position information, but also loses the ability to learn long-range dependencies. By comparing baseline and Row 1, the accuracy of the model without oversized convolution drops substantially by 0.6\% (79.4\% \emph{v.s.} 78.9\%) top-1 accuracy. It demonstrates that long-range dependencies are important to networks.

\noindent\textbf{Bifurcate gate units.} The bifurcate gate unit is an important mechanism to introduce data-driven operations into ParCNetV2. It increases the non-linearity and enhances the fitting ability. There is a degradation of 0.2\% (79.4\% \emph{v.s.} 79.2\%) without spatial BGU, and 0.3\%(79.4\% \emph{v.s.} 79.1\%) without channel BGU as shown in baseline, Row 2 and Row 3. It is similar to the data-driven operation of the squeeze-and-excitation block in ParC V1, while our BGU differs in the following two points. First BGU does not increase parameters. With $\Tilde{\alpha}=2.5$, our channel BGU is slightly more lightweight than the original FFN. Second, the two branches in our BGU are more balanced. They share a similar number of parameters and computational costs, unlike the heavy main branch and lightweight channel attention in most methods.

\noindent\textbf{Uniform local-global convolution.} The objective of the uniform local-global convolution block is to standardize the blocks used across various stages. In ParCNetv1, MobileNetV2 blocks had to be mixed with ParC blocks to construct the entire network. However, in ParCNet V2, the entire network is built by stacking ParCNet V2 blocks, as illustrated in Figure 1 in the supplementary material. This uniform design offers greater flexibility and ease of combination with other structures. Additionally, the uniform design results in a performance gain of 0.2\%.

\begin{table}
  \begin{center}
    \renewcommand\arraystretch{0.9}
    \setlength{\tabcolsep}{2.25pt}
    \small
    \begin{tabular}{l|ccccc}
      \toprule
      Models      & Param(M) & MACs & Latency$\downarrow$ & Memory$\downarrow$ & Top-1$\uparrow$ \\
                  & (M)      & (G)  & (ms)                & (MB)               & (\%)            \\
      \hline
      Swin-T      & 29       & 4.5  & 855                 & 139                & 81.3            \\
      ConvNeXt-T  & 29       & 4.5  & 875                 & 129                & 82.1            \\
      \rowcolor{cyan!10}
      ParCNetV2-T & 25       & 4.3  & \textbf{840}        & \textbf{118}       & \textbf{83.5}   \\
      \hline
      Swin-S      & 50       & 8.7  & 1576                & 222                & 83.0            \\
      ConvNeXt-S  & 50       & 8.7  & 1618                & 211                & 83.1            \\
      \rowcolor{cyan!10}
      ParCNetV2-S & 39       & 7.8  & \textbf{1485}       & \textbf{181}       & \textbf{84.3}   \\
      \hline
      Swin-B      & 88       & 15.4 & 2649                & 378                & 83.5            \\
      ConvNeXt-B  & 89       & 15.4 & 2708                & 364                & 83.8            \\
      \rowcolor{cyan!10}
      ParCNetV2-B & 56       & 12.5 & \textbf{2339}       & \textbf{252}       & \textbf{84.6}   \\
      \bottomrule
    \end{tabular}
  \end{center}
  \caption{\textbf{Inference on Arm (Quad Core Cortex-A17).} We compare the latency and memory cost during inference together with ImageNet-1K top-1 accuracy. Results are measured using RK3288 with batch size 1 and averaged over 100 iterations.}
  \label{tab:arm}
  \vspace{-3mm}
\end{table}

\subsection{Latency analysis}
\label{subsec:latency}
We analyze the inference latency of our ParCNetV2 on RTX3090 GPU and edge device RK3288. The Rockchip RK3288 is widely used in real-world applications such as smart TV and AI entrance guard system.

\noindent\textbf{GPU inference latency.} To ensure a fair comparison with large kernel convolution networks which use the \textit{implicit gemm} acceleration algorithm, such as RepLKNet~\cite{ding2022scaling} and SLaK~\cite{liu2022more}, we measure the inference latency of our ParCNetV2 models using a single NVIDIA RTX 3090 GPU with a batch size of 32, following the consistent implementation as theirs. As illustrated in Fig.~\ref{fig:trade_offs}, ParCNetV2 models achieve superior latency-accuracy trade-offs among large kernel networks, outperforming both Swin and ConvNeXt.

\noindent\textbf{Arm inference latency.} On RK3288, we port the models to the chip through ONNX and MNN and conducted each test for 100 iterations to measure the average inference speed. Tab.~\ref{tab:arm} demonstrates that ParCNetV2 runs faster and performs substantially better than Swin and ConvNeXt. Moreover, our model requires less memory, making it a more suitable option for edge computing applications.

\section{Conclusion}

This paper presents ParCNetV2, a pure convolutional neural network with state-of-the-art performance. It extends position-aware circular convolution with oversized convolutions and strengthens attention through bifurcate gate units. Besides, it utilizes a uniform local-global convolution block to unify the design of the early and late stage convolution blocks. We conduct extensive experiments on image classification and semantic segmentation to show the effectiveness and superiority of the proposed ParCNetV2 architecture.

\section*{Appendix}
\setcounter{section}{0}
\renewcommand\thesubsection{\Alph{subsection}}
\subsection{Introduction}
In this chapter, we present additional materials and results. First, we show some analysis of the model details. We present the proof that alternating the order of vertical and horizontal convolution does not affect the results of oversized convolution in Sec.~\ref{sec:commutative}. In Sec.~\ref{sec:alpha}, we explain how we adjust $\Tilde{\alpha}$ to fit the model size close to the original FFN. We also compare ParCNetV2 framework with the ParCNetV1 to show the simplicity of our model in Sec.~\ref{sec:v1vsv2}.

Then, we provide additional experiments analysis. In Sec.~\ref{sec:downstream}, we evaluate the performance of ParCNetV2 in object detection and semantic segmentation tasks, comparing it to other recently proposed models across various model scales. We show how we accelerate the inference with implicit gemm algorithm in Sec.~\ref{sec:inference}.

Finally, we show multiple visualization examples of the proposed ParCNetV2. On the one hand, We provide the corresponding standard convolution kernel of the separated oversized convolution, as well as a more detailed study of the proposed oversized convolution in Sec.~\ref{sec:visualize_kernel}. On the other hand, the comparison of Grad-CAM between the common convolution networks and ParCNetV2 is shown in Sec.~\ref{sec:gradcam}.

\subsection{Proof of the Commutative Property of Oversized Convoluiton}
\label{sec:commutative}
As mentioned in the paper, to compute the output of the oversized convolution $Z_{i,j}$ at location $(i, j)$, we use the following equations:
\begin{align}
   Y_{i,j} & = \sum_{s=-(H - 1)}^{H - 1}k^h_s X_{i+s, j}, \\
   Z_{i,j} & = \sum_{t=-(W - 1)}^{W - 1}k^w_t Y_{i, j+t}.
\end{align}
We combine the two equations and calculate $Z_{i,j}$ with a single function:
\begin{align*}
   Z_{i,j} & = \sum_{t=-(W - 1)}^{W - 1}k^w_t Y_{i, j+t}                                     \\
           & = \sum_{t=-(W - 1)}^{W - 1}k^w_t \sum_{s=-(H - 1)}^{H - 1}k^h_s X_{i+s, j + t}  \\
           & = \sum_{t=-(W - 1)}^{W - 1}\sum_{s=-(H - 1)}^{H - 1}k^w_t k^h_s X_{i+s, j + t}. \\
\end{align*}
Thus the separated oversized convolution can be regarded as a low-rank decomposition of a large convolution kernel ($k^h k^w$).
In addition, the commutative law of summation indicates that the order of addition does not influence the result. Thus the order of vertical and horizontal convolution does not affect the results of oversized convolution.

\subsection{Adjusting $\Tilde{\alpha}$ of Channel BGU}
\label{sec:alpha}
We adjust $\Tilde{\alpha}$ to fit the model size close to the original FFN. The number of parameters in the original FFN is $2\alpha C^2$, and in our FFN with BGU it is $2\Tilde{\alpha}C^2 + \Tilde{\alpha}C^2=3\Tilde{\alpha}C^2$. To keep the number of parameters almost unchanged, we get $2\alpha C^2=3\Tilde{\alpha}C^2$, thus
\begin{equation}
   \Tilde{\alpha}=2\alpha/3.\label{eq:expansion_ratio}
\end{equation}
The expanded ratio of FFN in most existing models is $4$, which indicates that $\Tilde{\alpha} = 8 / 3$. Researchers have shown that when the number of channels is a multiple of 32, it is beneficial for hardware optimization~\cite{cuda}, so we choose $\Tilde{\alpha} = 2.5$ to approximate the original FFN.

\subsection{Comparison ParCNetV2 and ParCNetV1 Framework}
\label{sec:v1vsv2}

\begin{figure*}
    \centering
    \includegraphics[width=0.9\linewidth]{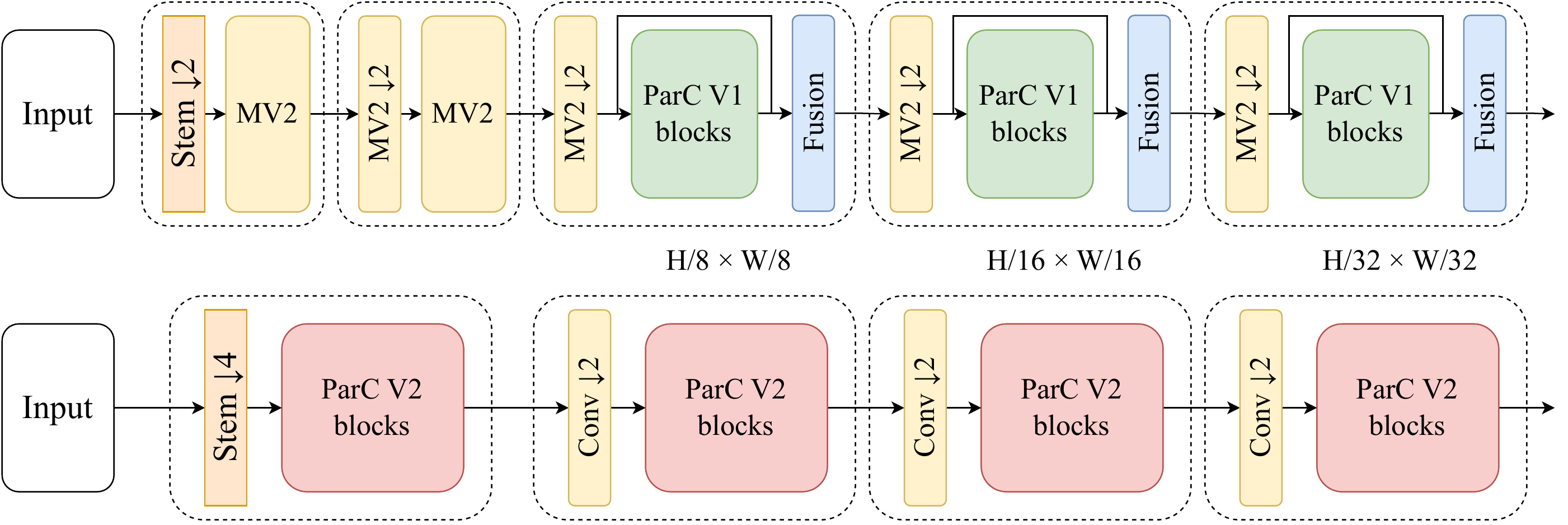}
    \caption{\textbf{Framework comparison between ParCNetV1 and ParCNetV2.} Downsampling modules with downsampling ratio 2 and 4 are represented by $\downarrow 2$ and $\downarrow 4$, respectively. \textbf{MV2}: MobileNetV2 block.}
    \label{fig:v1vsv2}
\end{figure*}

We compare the framework of ParCNetV1 and ParCNetV2 in Fig.~\ref{fig:v1vsv2}. ParCNetV1 is a complicated model with multi-branch architecture. The fusion modules are necessary to combine local features from MobileNetV2 block and ParC V1 block. While in our ParCNetV2, the whole model utilizes the same ParC V2 blocks. Our method is easy to follow, and consistent to the widely-used 4-stage framework.

\subsection{Additional Experiments on Downstream Tasks}
\label{sec:downstream}
\begin{table}
   \centering
   \setlength{\tabcolsep}{1pt}
   \small
   \begin{tabular}{l|cccccc}
      \toprule
      backbone    & AP$^{bbox}$                                              & AP$^{bbox}_{50}$ & AP$^{bbox}_{75}$ & AP$^{mask}$   & AP$^{mask}_{50}$ & AP$^{mask}_{75}$ \\
      \hline
                  & \multicolumn{6}{c}{Mask R-CNN $3\times$ schedule}                                                                                                     \\
      Swin-T      & 46.0                                                     & 68.1             & 50.3             & 41.6          & 65.1             & 44.9             \\
      ConvNeXt-T  & 46.2                                                     & 67.9             & 50.8             & 41.7          & 65.0             & 44.9             \\
      \rowcolor{cyan!10}
      ParCNetV2-T & \textbf{48.9}                                            & 70.3             & 53.9             & \textbf{43.7} & 67.6             & 47.0             \\
      \hline
                  & \multicolumn{6}{c}{Cascade Mask R-CNN $3\times$ schedule}                                                                                             \\
      Swin-T      & 50.4                                                     & 69.2             & 54.7             & 43.7          & 66.6             & 47.3             \\
      ConvNeXt-T  & 50.4                                                     & 69.1             & 54.8             & 43.7          & 66.5             & 47.3             \\
      \rowcolor{cyan!10}
      ParCNetV2-T & \textbf{52.6}                                            & 71.0             & 57.3             & \textbf{45.6} & 68.6             & 49.8             \\
      \hdashline
      Swin-S      & 51.9                                                     & 70.7             & 56.3             & 45.0          & 68.2             & 48.8             \\
      ConvNeXt-S  & 51.9                                                     & 70.8             & 56.5             & 45.0          & 68.2             & 48.8             \\
      \rowcolor{cyan!10}
      ParCNetV2-S & \textbf{53.4}                                            & 72.1             & 58.4             & \textbf{46.3} & 69.6             & 50.2             \\
      \hdashline
      Swin-B      & 51.9                                                     & 70.5             & 56.4             & 45.0          & 68.1             & 48.9             \\
      ConvNeXt-B  & 52.7                                                     & 71.3             & 57.2             & 45.6          & 68.9             & 49.5             \\
      \rowcolor{cyan!10}
      ParCNetV2-B & \textbf{54.0}                                            & 72.6             & 58.6             & \textbf{46.7} & 70.2             & 51.1             \\
      \bottomrule
   \end{tabular}
   \caption{Comparisons on \textbf{COCO~\cite{lin2014microsoft} object detection and instance segmentation.} We use Mask R-CNN~\cite{he2017mask} and Cascade Mask R-CNN~\cite{cai2018cascade} as a basic framework. All models are pretrained on ImageNet-1K and trained on COCO for $3\times$ iterations.}
   \label{tab:full_detection}
   \vspace{-0mm}
\end{table}

\noindent\textbf{Object detection and instance segmentaion on COCO.}
Following previous works~\cite{liu2021swin,liu2022convnet}, we finetune Cascade Mask R-CNN~\cite{cai2018cascade} on COCO dataset~\cite{lin2014microsoft} with ParCNetV2 backbones. MMDetection~\cite{chen2019mmdetection} is used as the base framework. All models use pre-trained weights from ImageNet1K and are trained with $3\times$ schedule with multi-scale training. The experiment settings follow~\cite{liu2022convnet}.
We follow all the experiment settings of ConvNeXt~\cite{liu2022convnet} except that the number of layers in layerwise learning rate decay~\cite{bao2021beit} are adjusted to $\{7, 13, 13\}$ to fit with our model.
Tab.~\ref{tab:full_detection} shows object detection and instance segmentation results comparing our ParCNetV2 with Swin~\cite{liu2021swin} and ConvNeXt~\cite{liu2022convnet}. ParCNetV2 outperforms both the transformer network and convolution network by a large margin across different model complexities.

\begin{table}
   \centering
   \setlength{\tabcolsep}{6pt}
   \small
   \begin{tabular}{l|cccc}
      \toprule
      backbone     & Param & FLOPs & mIoU ss       & mIoU ms       \\
                   & (M)   & (G)   & (\%)          & (\%)          \\
      \hline
      Swin-T       & 60    & 945   & -             & 45.8          \\
      ConvNeXt-T   & 60    & 939   & 46.0          & 46.7          \\
      SLaK-T       & 65    & 936   & 47.6          & -             \\
      \rowcolor{cyan!10}
      ParCNetV2-T  & 55    & 932   & \textbf{48.5} & \textbf{49.4} \\
      \hline
      Swin-S       & 81    & 1038  & -             & 49.5          \\
      ConvNeXt-S   & 82    & 1027  & 48.7          & 49.6          \\
      SLaK-S       & 91    & 1028  & 49.4          & -             \\
      \rowcolor{cyan!10}
      ParCNetV2-S  & 69    & 1005  & \textbf{50.0} & \textbf{51.0} \\
      \hline
      Swin-B       & 121   & 1188  & 48.1          & 49.7          \\
      ConvNeXt-B   & 122   & 1170  & 49.1          & 49.9          \\
      RepLKNet-31B & 112   & 1170  & 49.9          & 50.6          \\
      SLaK-B       & 135   & 1172  & 50.2          & -             \\
      \rowcolor{cyan!10}
      ParCNetV2-B  & 87    & 1105  & \textbf{50.2} & \textbf{51.1} \\
      \bottomrule
   \end{tabular}
   \caption{Comparisons on \textbf{ADE20K~\cite{zhou2019semantic} semantic segmentation.} We use UperNet as a basic framework. All models are pretrained on ImageNet-1K and trained on ADE20K for 160K iterations. FLOPs are measured with the input size of (2048, 512). \textbf{ss} and \textbf{ms} indicates single-scale and multi-scale testing, respectively.} 
   \label{tab:full_segmentation}
   \vspace{-3mm}
\end{table}

\begin{figure*}
    \centering
    \includegraphics[width=0.9\linewidth]{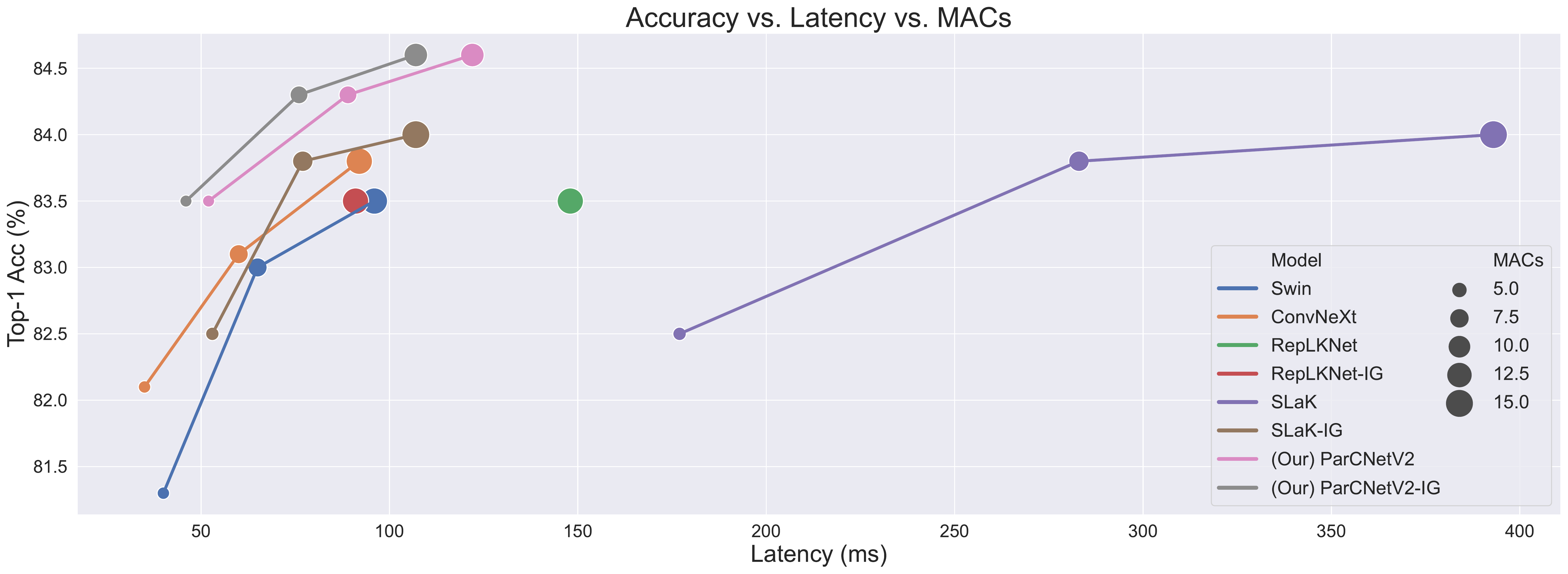}
    \caption{\textbf{Inference time and model accuracy.} \textbf{IG}: implicit gemm acceleration.}
    \label{fig:inference}
\end{figure*}

\noindent\textbf{Semantic segmentation on ADE20K.}
ADE20K~\cite{zhou2019semantic} is a widely-used semantic segmentation dataset, covering a broad range of 150 semantic categories. It has 25K images in total, with 20K for training, 2K for validation, and another 3K for testing. In this paper, we trained our ParCNetV2 on the training set, and report mIoUs
on the validation set with both single-scale testing and multi-scale testing.

\begin{figure}
   \centering
   \begin{subfigure}{0.48\linewidth}
      \centering
      \includegraphics[height=\linewidth]{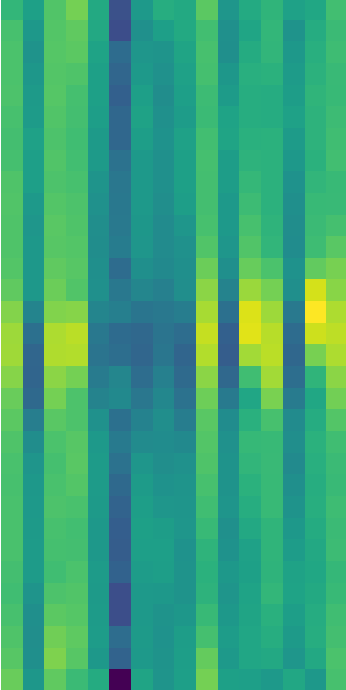}
      \caption{Vertical convolution kernels. Each column is a channel of the vertical kernel.}
   \end{subfigure}
   \begin{subfigure}{0.48\linewidth}
      \centering
      \includegraphics[width=\linewidth]{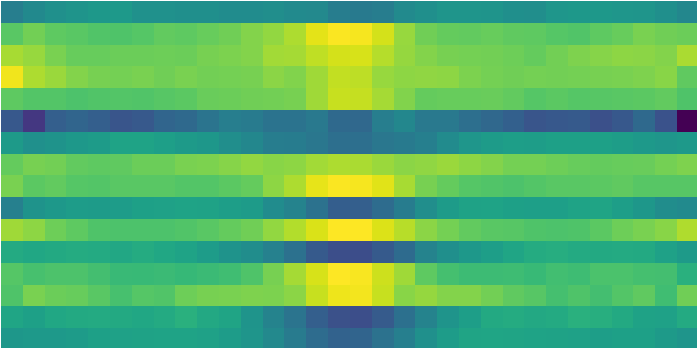}
      \caption{Horizontal convolution kernels. Each row is a channel of the horizontal kernel.}
   \end{subfigure}
   \caption{The vertical and horizontal oversized convolution kernel of the last uniform block of the third stage. We randomly selected 16 channels as examples.}
   \label{fig:separated_oversized}
\end{figure}

We finetune UperNet~\cite{xiao2018unified} in mmsegmentation as our base framework. 
Following Swin~\cite{liu2021swin} and ConvNeXt~\cite{liu2022convnet} settings in training, 
we employ the AdamW~\cite{kingma2014adam} optimizer with an initial learning rate of $1\times10^{-4}$.
We use stage-wise learning rate decay~\cite{bao2021beit} as ConvNeXt.
We also employ a linear warmup of 1500 iterations with initial learning rate $1\times10^{-6}$.
We adjust the weight decay to $0.02$.
All models use pre-trained weights from ImageNet1K and are trained on 8 GPUs with 2 images per GPU for 160K iterations.
For augmentations, we adopt the default setting in mmsegmentation of random horizontal flipping, random re-scaling within ratio range [0.5, 2.0] and random photometric distortion.
Stochastic depth with ratio for ParCNetV2-T, ParCNetV2-S, ParCNetV2-B are set to 0.3, 0.3, and 0.5, respectively.
All the models are trained on the standard setting as the previous approaches with an input of 512$\times$512.
Tab.~\ref{tab:segmentation} lists the model size, FLOPs, and mIoU of single-scale and multi-scale testing for different backbones.

\begin{figure}
   \centering
   \includegraphics[width=0.96\linewidth]{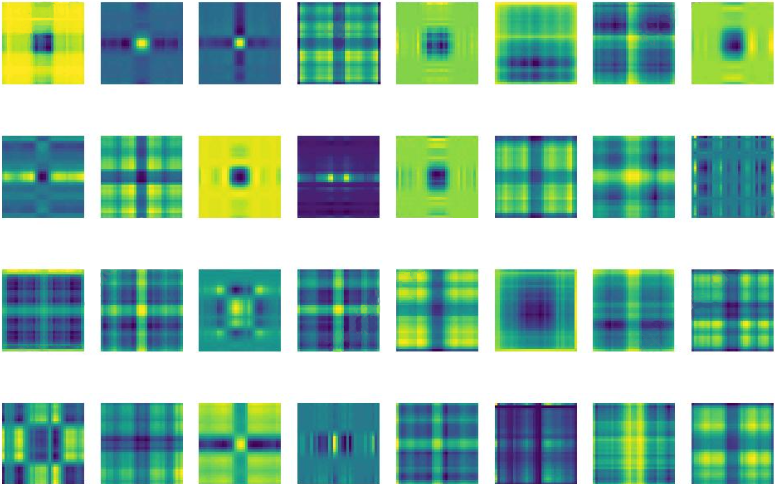}
   \caption{The corresponding oversized convolution kernel of the last uniform block of the third stage. We randomly selected 32 channels as examples.}
   \label{fig:reconstructed_standard}
\end{figure}

\subsection{Inference Acceleration}
\label{sec:inference}

\begin{figure*}
   \centering
   \includegraphics[width=\linewidth]{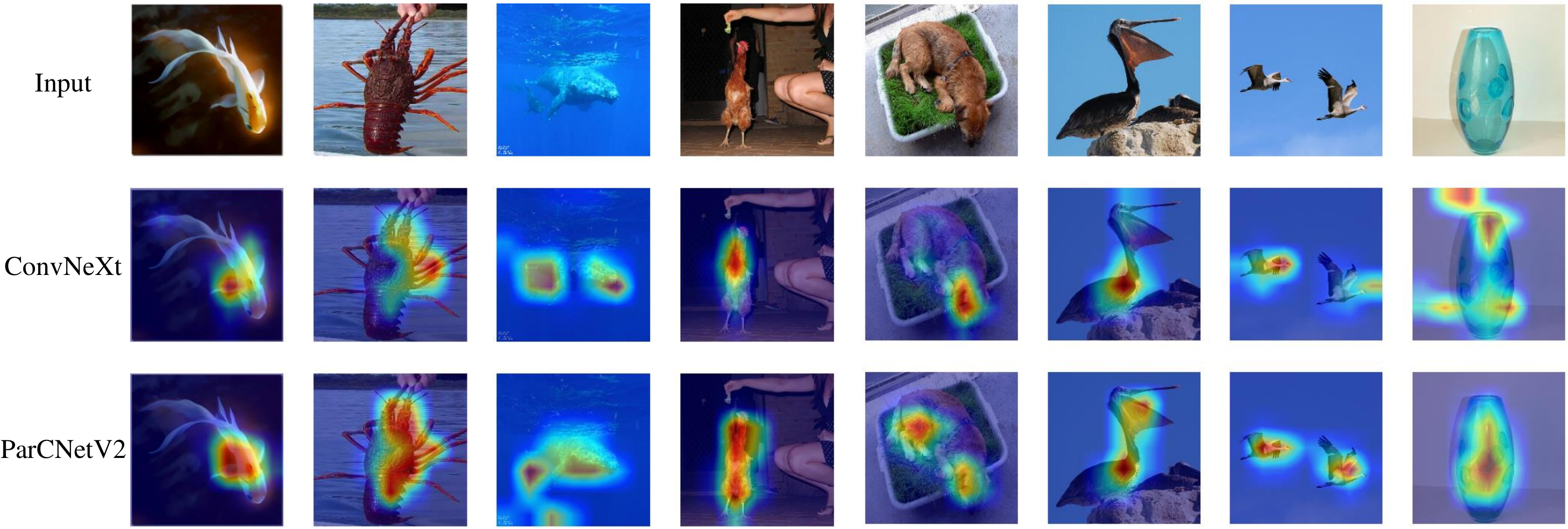}
   \caption{The Grad-CAM of ConvNeXt and our proposed ParCNetV2. The first line is the original image, the second line is the Grad-CAM for ConvNeXt, and the third line is our ParCNetV2.}
   \label{fig:grad_cam}
\end{figure*}

We implement the implicit gemm algorithm as~\cite{ding2022scaling}. To speed up ParCNetV2, we first reconstruct the standard convolution kernel with reparameterization, including separative oversized convolution and local $7\times7$ convolution. Then we use implicit gemm algorithm to implement the depthwise convolution. It is worth noting that this transform brings a bit more computational complexity, and only the convolutions of the last three stages run faster under these operations.

Tab.~\ref{fig:inference} show the original and accelerated inference time of ParCNetV2. As illustrated in Figure~\ref{fig:inference}, our proposed ParCNetV2 benefits from optimized algorithms. However, it does not heavily rely on optimization. Even without optimization, ParCNetV2 achieves a better balance between accuracy and speed compared to other large kernel models that have been optimized, such as RepLKNet~\cite{ding2022scaling} and SLaK~\cite{liu2022more}. However, dropping specific optimization for other large kernel models, especially SLaK, significantly affects their speed (as shown by the transition from the earth-colored line to the purple line). After optimization, ParCNetV2 exhibits clear advantages.

\subsection{Visualization of Local and Oversized Convolutions}
\label{sec:visualize_kernel}
Our proposed ParCNet V2 involves using an oversized convolution kernel with dimensions $C\times (2H - 1)\times 1$ and $C\times 1\times (2W - 1)$, as illustrated in Fig.\ref{fig:separated_oversized}. This oversized kernel is effective in capturing global context with a smoother kernel. For further analysis, we reconstruct a sequence of vertical and horizontal convolution kernels into 2D convolution kernels, as shown in Figure\ref{fig:reconstructed_standard}. We observe that different kernels have distinct characteristics, with some focusing on local features and others on longer-range features. This behavior is similar to the attention maps used in vision transformers~\cite{dosovitskiy2020image,raghu2021vision}. Viewed in 2D, the oversized convolution kernels exhibit a wide range of diversity, which makes them well-suited for handling complex global contexts.

\subsection{Visualization of Grad-CAM}
\label{sec:gradcam}

We compare the Grad-CAM~\cite{selvaraju2017grad} of our ParCNetV2 against the strong baseline ConvNeXt~\cite{liu2022convnet}. ParCNetV2 utilizes global oversized convolutions and an attention mechanism of bifurcate gate units. As shown in Fig.~\ref{fig:grad_cam}, ParCNetV2 either focuses on larger areas of the objects or produces a more smooth activation map, which indicates that our model has a stronger ability to capture large objects and texture features.

  {\small
    \bibliographystyle{ieee_fullname}
    \bibliography{egbib}
  }

\end{document}